\def\year{2021}\relax
\documentclass[letterpaper]{article} % DO NOT CHANGE THIS
\usepackage{aaai21}  % DO NOT CHANGE THIS
\usepackage{times}  % DO NOT CHANGE THIS
\usepackage{helvet} % DO NOT CHANGE THIS
\usepackage{courier}  % DO NOT CHANGE THIS
\usepackage[hyphens]{url}  % DO NOT CHANGE THIS
\usepackage{graphicx} % DO NOT CHANGE THIS
\usepackage{natbib}  % DO NOT CHANGE THIS AND DO NOT ADD ANY OPTIONS TO IT
\usepackage{caption} % DO NOT CHANGE THIS AND DO NOT ADD ANY OPTIONS TO IT
\usepackage{booktabs}
\usepackage{amsmath}
\usepackage{xspace}
\usepackage{multirow}
\usepackage[dvipsnames]{xcolor}
\usepackage{algorithmic}
\usepackage[ruled,vlined]{algorithm2e}
\usepackage{multirow}

\newcommand{\natcat}{\textsc{NatCat}\xspace}
\newcommand{\kevin}[1]{\textcolor{blue}{\bf \small [ #1 --KPG]}}
\newcommand{\karl}[1]{\textcolor{brown}{\bf \small [ #1 --Karl]}}
\newcommand{\kpgcomment}[1]{\textcolor{blue}{\bf \small [ #1 --KPG]}}
\newcommand{\zewei}[1]{\textcolor{red}{\bf \small [ #1 --Zewei]}}

\renewcommand{\kevin}[1]{}
\renewcommand{\zewei}[1]{}
\renewcommand{\karl}[1]{}
\renewcommand{\kpgcomment}[1]{}

\newenvironment{itemizesquish}{\begin{list}{\setcounter{enumi}{0}\labelitemi}{\setlength{\itemsep}{-0.25em}\setlength{\labelwidth}{0.5em}\setlength{\leftmargin}{\labelwidth}\addtolength{\leftmargin}{\labelsep}}}{\end{list}}

\def\wa{\textsuperscript{\rm 1}}
\def\wb{\textsuperscript{\rm 2}}
\def\wc{\textsuperscript{\rm 3}}

\newcommand{\twentynews}{\textsc{20 Newsgroups}\xspace}
\newcommand{\esa}{\textsc{Explicit Semantic Analysis}\xspace}
\newcommand{\bert}{\textsc{BERT}\xspace}
\newcommand{\roberta}{\textsc{RoBERTa}\xspace}

\newcommand{\agnews}{AG\xspace}
\newcommand{\dbp}{DBP\xspace}
\newcommand{\yahoo}{Yahoo\xspace}
\newcommand{\ulr}{ULR\xspace}
\newcommand{\enc}{\mathrm{enc}}

\title{Self Training in Text Classification}
\title{K-Means Can Improve Zero-shot Text Classifiers}
\title{Semi-supervised K-Means Clustering for Text Classification}
\title{Unsupervised Refinement Improves Dataless Text Classification}
\title{Unsupervised Label Refinement Improves Dataless Text Classification}
\author{\textbf{Zewei Chu},\wa \xspace \Large \textbf{
Karl Stratos},\wb \xspace \Large \textbf{Kevin Gimpel}\wc \\
}
\affiliations{
\wa The University of Chicago, 5730 S Ellis Ave, Chicago, IL 60637, USA\\
\wb Rutgers University, 110 Frelinghuysen Road Piscataway, NJ 08854, USA\\
\wc Toyota Technological Institute at Chicago, 6045 S Kenwood Ave, Chicago, IL 60637, USA\\
zeweichu@gmail.com, karlstratos@gmail.com, kgimpel@ttic.edu 
}
\begin{document}

\maketitle

\begin{abstract}

Dataless text classification is capable of classifying documents into previously unseen labels by assigning a score to any document paired with a label description. While promising, it crucially relies on accurate descriptions of the label set for each downstream task. This reliance causes dataless classifiers to be highly sensitive to the choice of label descriptions and hinders the broader application of dataless classification in practice. In this paper, we ask the following question: how can we improve dataless text classification using the inputs of the downstream task dataset? Our primary solution is a clustering based approach. Given a dataless classifier, our approach refines its set of predictions using $k$-means clustering. We demonstrate the broad applicability of our approach by improving the performance of two widely used classifier architectures, one that encodes text-category pairs with two independent encoders and one with a single joint encoder. Experiments show that our approach consistently improves dataless classification across different datasets and makes the classifier more robust to the choice of label descriptions. \footnote{Code and data available at \url{https://github.com/ZeweiChu/ULR}. }

\end{abstract}

\section{Introduction}
\label{sec:intro}

\begin{table*}[t]
\setlength{\tabcolsep}{4pt}
\small
\begin{center}
\begin{tabular}{l|cccccc}
\toprule
 & \bf world & international & world politics & world news & world & world news \\
choice of & \bf sports & health & sports & health & health and sports & sports \\
label names & \bf business & finance & business and finance & business & commerce & finance \\
& \bf science technology & technology & science and technology & science and technology & science technology & science \\
\midrule
dataless & 69.9 &  55.9 & 71.6 & 49.9 & 55.0 & 68.8 \\
dataless + \ulr & 70.7 & 78.3 & 78.4 & 70.2 & 70.9 & 76.1 \\
% \midrule
% supervised training & \multicolumn{6}{c}{92.4} \\
% Human & \multicolumn{6}{c}{83.8} \\
\bottomrule
\end{tabular}
\end{center}
\caption{\label{tab:example-noisy-labels} Accuracy (\%) of a \roberta dual encoder dataless classifier\footnote{details in Section~\ref{sec:experiments}} on AG News with different choices of label names. 
% An example of how variations on choices of label names lead to different zero-shot text classification model performances. 
% The experiment is conducted on AGNews, a fourway text classification task on a news dataset. 
The original label set (boldfaced) is ``world'', ``sports'', ``business'', and ``sci/tech''. 
%``dataless'' is the performance, and 
The ``dataless + \ulr'' row shows accuracies after applying unsupervised label refinement (details in Section~\ref{sec:ulr}). 
%\footnote{details in Section~\ref{sec:ulr}}.
% We use a \roberta based dual encoder model as the dataless classifier. 
%\zewei{Is this ``human'' accurcy necessary? Or shall we just remove it. }
%\karl{I don't think it's necessary. But explain what dataless+ULR is (e.g., ``dataless+ULR is the dataless classifier with the unsupervised label refinement (ULR) technique in Section 3'').}
%\zewei{I have updated the example results with the L2 trained \roberta dual encoder}
%\karl{Cool, this makes the point much more clear.}
%\kevin{great! I'm wondering if we should switch out one or two of the sets with ``health'' for other sets that use sports instead. it seems like all the ones with ``health'' are doing badly, and health is one that we added intentionally to be noisy, so it's not really what a researcher would do in the real world. Is there a set that uses sports instead of health that has accuracy in the 50s?} 
% \zewei{I changed the last column here}\kevin{great, thanks!}
}
\end{table*}

% In this paper, we show how weakly supervised text classification models can be improved by two simple tricks, self-training and k-means clustering. 
Dataless text classification aims at classifying text into categories without using any annotated training data from the task of interest. %same domain. 
Prior work~\citep{chang-dataless, song-dataless-hierarchical} has shown that with effective ways to represent texts and labels, dataless classifiers can perform text classification on unbounded label sets if suitable descriptions of the labels are provided. 

There have been many previous efforts in dataless or zero-shot text classification
% Some embed text and class labels into the same embedding space and use simple methods for classification 
\citep{dauphin2013zero,nam2016all, li-label-embedding-16,ma-label-embedding-16,shu2017doc,fei2016breaking,zhang-etal-2019-integrating-dupe,yogatama-generative-discrimitive,mullenbach2018explainable,rios2018few,meng2019weakly}.
Many different settings have been considered across this prior work, and some have used slightly different definitions of dataless classifiers. 
In this paper, we use the term ``dataless text classification'' to refer to methods that: (1) can assign scores to any document-category pair, and 
%for semantic relatedness, 
(2) do not require any annotated training data from downstream tasks. A dataless classifier can therefore be immediately adapted to a particular label set in a downstream task dataset by scoring each possible label for a document and returning the label with the highest score. 
Dataless classifiers are typically built from large-scale freely available text resources such as Wikipedia~\citep{chang-dataless, yin2019benchmarking}. 

% Prior work~\citep{yin2019benchmarking} show that by leveraging text and category resources from a variety of domains, dataless classifiers can be built to perform text classifications on unbounded sets of categories provided good label descriptions of a downstream task. \zewei{This sentence may be inaccurate. When I say ``dataless'', people may be thinking of \citet{chang-dataless}. Is it OK to refer to \cite{yin2019benchmarking} as ``dataless''? }

% \zewei{Following \citet{chang-dataless}, in this work by ``dataless'' text classification we mean methods that can assign scores to any document-category pairs for semantic relatedness, and do not require any training on downstream task datasets. }

%\kevin{I think that's a very clear problem with dataless classification of all kinds, especially dataless text classification, and I think suboptimal label descriptions are a big reason why supervised methods do so much better than dataless methods on these tasks. I think we could motivate that problem by mentioning early on in the paper the large variance we see with different label name sets as in your recent results. Then we can say that in this paper we ask the question: how can we improve dataless text classification by using the inputs of the downstream task datasets in addition to the set of label names? Our primary solution is clustering. (We could also include self-training in the experimental comparison as it's another way of using the inputs and set of label names, and it's a natural idea to try.) }

% However, in dataless text classification, i
A well known problem with dataless classifiers is that the choice of label names has a significant impact on performance~\citep{chang-dataless}. 
As dataless classifiers rely purely on the label descriptions in a downstream task, there is typically no tailoring or fine-tuning of the classifier for a given dataset. 
A poor choice of label descriptions could jeopardize  the performance of dataless classifiers on a particular text classification task, so prior work has addressed this with modifications to label descriptions.  \citet{chang-dataless} manually expand the label names of the 20 newsgroups dataset. \citet{yin2019benchmarking} expand  labels by their WordNet definitions. 
% For instance, ``sports'' is interpreted as ``an active diversion requiring physical exertion and competition''. \kevin{this last sentence could be removed if we need to save space}

To illustrate the problem, Table~\ref{tab:example-noisy-labels} shows various choices of label names when applying a dataless classifier to the 4-class AG News dataset. 
When we change the descriptions of the four labels, performance of our dataless text classifier\footnote{\roberta dual encoder architecture as described in Section~\ref{sec:experiments}} changes drastically. 
The broader application of %the seemingly promising 
dataless text classifiers is hindered by their fragility caused by the choice of label descriptions. 
It is unclear how practitioners should choose label descriptions for practical use.

In this paper, we ask the following question: how can we improve dataless text classification provided the \emph{unlabeled} set of input texts for the downstream task in addition to its label descriptions? 
Our approach, which we refer to as unsupervised label refinement (\ulr), is based on $k$-means clustering. 
We develop variations of our approach so that it can be applied to different styles of dataless text classifiers to improve their performance. 
Table~\ref{tab:example-noisy-labels} shows results when applying \ulr to our dataless text classifier. 
In all cases, accuracies improve after applying \ulr, with larger gains when using weaker label descriptions. 
%The gaps between dataless text classifiers and the supervised model are narrowed by \ulr. 
% and it is very close to the human performance. 

% \zewei{Let us give a motivating example as of how the choice of label names could affect the model performances. }

% This paper takes a step further to improve such zero-shot text classifiers by incorporating the knowledge from the domains of downstream tasks. Our proposed method is K-Means clustering on the downstream tasks. Our experimental results show that K-means almost always improve the model performances on a variety of text classification tasks. 

% In this paper, we propose a method based on K-Means clustering to mitigate the randomness brought by label names. 

% This also makes me wonder whether we would see similar improvements when starting with ESA in place of our transformer encoders, if we want to follow more closely upon the prior work in dataless classification. Or, if not, we could just say that our methods are better than ESA across datasets so we just focus on refinement with our models. 

% We propose XXXX algorithm. It is an algorithm of 

Our contributions in this paper are:
\begin{itemizesquish}
\item We propose unsupervised label refinement (\ulr), a $k$-means clustering based approach to improve dataless classifiers. 
\item 
% ULR can be applied to both dual encoder or single encoder architectures of dataless classifiers. Experiments show that ULR almost always improves the performances of such dataless classifiers. \zewei{
We develop variations of \ulr that can be applied to different model architectures of dataless classifiers. Experiments on dual encoder and single encoder architectures show that \ulr almost always improves performance. 
% }
\item Experiments show that \ulr improves robustness of dataless classification against choices of label names, making dataless classifiers more practically useful. 
\end{itemizesquish}

% \section{XXXX algorithm}

% \zewei{In this section, I want to give an overview of our proposed method for zero-shot text classification (again I don't know whether zero-shot is a good terminology. )}

% XXXX algorithm is composed of two stages: zero-shot text classifier training, and K-Means text clustering. 

% \zewei{Put a graph here showing how our proposed approach works. }
% \karl{Present the main ideas in the first three pages}

\section{Background: Dataless Text Classification}
\label{sec:background}

% \zewei{This is the first step of our XXXX framework. Our goal is to train an imperfect general purpose text classifier. }
%\kevin{This section talks about text classification, but not about \emph{dataless} text classification. I think we should state here the key characteristics of dataless text classification, how it assumes the ability to compute scores for an unbounded set of categories and is typically evaluated by using a single dataless text classifier for a set of downstream tasks with different label sets. There is typically no tailoring or fine-tuning of the classifier for a dataset other than through specifying the label descriptions. Also might be good to give some examples and citations of dataless text classifier architectures from past work?}
%\zewei{I have rewritten this following part}

Dataless text classification \citep{chang-dataless, chen2015dataless, song-dataless-hierarchical, yin2019benchmarking} aims at building a single, universal text classifier that can be applied to any text classification task with a given set of label descriptions. Dataless classifiers can be used on an unbounded set of categories. There is typically no tailoring or fine-tuning of the classifier for a dataset other than through specifying the label descriptions.

Since annotated data in the target task is not available for training,
% dataless classification is sensitive to the choice of label descriptions. 
the choice of label descriptions plays a critical role in the performance of dataless classifiers
~\citep{chang-dataless,yin2019benchmarking}. 
%\kevin{feels like a citation is needed here to a paper that shows the accuracies with different choices of label descriptions, or at least does label expansion} 
With a dataless classifier, a score is produced for each text-category pair, indicating their semantic relatedness. Text classification then becomes a ranking problem, i.e., picking the category that has the highest semantic relatedness with the text.

%\kevin{using citet with a list of citations leads to weird phrasing. can you change it so that it reads more naturally?} \zewei{Does citep look better? }\kevin{I moved the cites to the end.. i think it's more natural this way }
Several have used \esa (ESA) \citep{gabrilovich-esa} as text representations in dataless text classification \citep{chang-dataless,song-dataless-hierarchical,wang-universal-2009}. 
% In their work, they manually expand the label names of the 20 news groups dataset so they capture the semantics of each category. 
Both label descriptions and text are encoded into ESA vectors. Cosine similarity is used to compute scores between text and categories. \citet{yin2019benchmarking} directly compute text-category relatedness with a single \bert~\citep{devlin2018bert} model. 
% Most dataless classifiers require good descriptions of labels. 
% \citet{chang-dataless} uses Explicit Semantic Analysis (ESA)~\citep{gabrilovich-esa} as text and category representations.
% Text classification aims at assigning a given text to its most suitable class among a set of categories. For a particular text classification task, let the set of all possible classes be $\mathcal{C}$. Given a piece of text $t$, and a scoring function $\mathrm{score}$ that measures how likely the text belongs to a particular category $c \in \mathcal{C}$, we now have a score for each category $\mathrm{score}(c, t) \forall c \in C$. 
%Text classification can be treated as a ranking problem. \kevin{commented this out because i'm not really sure if i agree with this.. there's only one correct class so any ranking that is consistent with the correct class having the top rank is equally good; we don't care about the ranks of the other classes. it's still just a classification problem.}
% The text is categorized into the class with the highest score $\max_{c \in C}\mathrm{score}(c, t)$. 
\citet{chang-dataless} and \citet{yin2019benchmarking} exemplify two typical modeling choices for dataless classifiers, namely dual encoder and single encoder architectures, respectively. We will introduce them briefly and consider both types in our experiments. 

\paragraph{Dual encoder model.} 
\begin{figure}[t]
    \centering
    \includegraphics[scale=0.45]{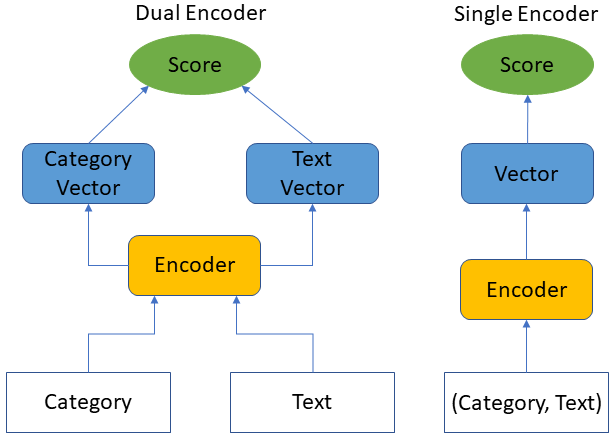}
    \caption{Dual encoder and single encoder architectures}
    \label{fig:encoders}
\end{figure}

With the dual encoder model, the category and text are fed into the encoder separately, each producing a vector representation. 
The text and category encoders could have shared or independent parameters. In our experiments, we always share parameters, i.e., we use the same encoder for both the categories and texts. 
% , i.e., the two encoders are identical.  
% are produced for them respectively. 
A distance function takes both the category and text vectors and produces a scalar value. 
In our experiments, this scoring function can be either cosine distance or Euclidean (L2) distance. 

% \kevin{Since we always share parameters in our experiments, we should say something like the following:} \zewei{Looks great} 

\paragraph{Single encoder model.}
With a single encoder model, the category is combined with the text as a single sequence and fed into an encoder. The output of the encoder is a single vector that contains the information from both the category and the text. This vector can pass through a linear layer and produce a score for this particular document-category pair.
% \begin{figure}[t]
%     \centering
%     \includegraphics[scale=0.45]{images/monoencoder.png}
%     \caption{single encoder architecture 
%     }
%     \label{fig:single_encoder}
% \end{figure}
% Figure~\ref{fig:single_encoder} demonstrates the architecture of a typical single encoder model for text classification. 

Figure~\ref{fig:encoders} demonstrates the architectures of typical dual and single encoder models for text classification.

% \paragraph{Prediction}
% At prediction time, either a single encoder model or a dual encoder model can be applied on a particular text classification task with a fixed set of labels. The task can be performed as a ranking problem. More specifically, given some text to be categorized, we score the text with all potential categories. The text is classified into the category with the highest score among them. 

\section{Unsupervised Label Refinement (ULR)}
\label{sec:ulr}
%\zewei{The name ``unsupervised label refinement'' may be inappropriate here, since in the single encoder model, we directly refine the scores, i.e., both the text and labels instead of only the labels. }
%\kevin{Hmmm.. my reading of Alg 2 suggests that the $P_k$'s never change, rather only the $ct_i$'s change, which are a stand-in for label embeddings. Or have I misunderstood?}
%\zewei{You are right. I was }
% \zewei{This is the second component of our proposed XXXX framework. By leveraging the unlabeled text from the prediction domain, our K-Means algorithm could improve the model performance. }

In this section, we introduce unsupervised label refinement (\ulr). \ulr uses the components of a dataless classifier and refines representations of labels with a modified $k$-means clustering algorithm. 
% The intuition is that supervised text classification can be combined with unsupervised text clustering. 
%\kevin{I think we probably don't want to use the term ``supervised'' as that may confuse the reader. I think we can state the intuition with something like the following:} \zewei{agreed and removed this sentence} 
While dataless text classifiers are designed to handle an unbounded set of categories, they are used and evaluated on a particular set of documents with a set of labels. The idea of our approach is to leverage the assumption that the documents in a text classification dataset are separable according to the accompanying set of labels. That is, given a strong document encoder, the documents should be separable by label in the encoded space. This assumption is similarly made when performing clustering for unsupervised document classification~\citep{liang2009online}.  
%\kevin{paper to cite here? at a minimum, we could cite Percy Liang's stepwise EM paper, which I think used EM for unsup document classification}

We use the set of unlabeled input texts to refine the predictions of our dataless classifiers via clustering. To better inform the algorithm, we initialize the clusters by using our dataless classifiers run on the provided label set for each task. The algorithm takes on different forms for the dual and single encoder models. Details are provided below.

\subsection{\ulr for Dual Encoder Architectures}
In the setting of a dual encoder model, 
% with Euclidean distance as a scoring function
% , we propose to apply $k$-means clustering when applying this model to classify a set of text (documents). 
we propose to perform $k$-means clustering among text representations, i.e., of vectors produced by the text encoder. 
% The hope 
The assumption
is that texts falling under the same category will be close in the semantic space. 
% The hope is that texts falling under the same category will be close in the semantic space. 
%\kevin{i think this is more an assumption than a hope, right? i think the intuition is something more like: we want to adapt our dataless classifier's predictions based on the natural clustering structure in the encodings of the texts} \zewei{updated above}
We want to adapt our dataless classifier's predictions based on the natural clustering structure in the encodings of the texts. 

% The intuition of applying K-means in the dual-encoder setting is that the representations of text falling under the same category are deemed to be close with each other. 
% \zewei{Is this claim right? }
% The task of text classification in such scenario can be treated as a problem of clustering. The key is to find the ``centroids'' of each category of the label set. 

% In the setting of a dual encoder model, 
% the category vector produced by the encoder can be naturally treated as the initial centroid.  

% \kevin{Write down the difference between original $k$-means. and early stopping. }

% In this algorithm, the centroids of the clusters are initialized as the encoded category names. 
% The text is also encoded to become vectors. A scoring function (in our work, either cosine similarity or euclidean distance) is applied to all combinations of category and text vectors, At every iteration, each text is re-assigned to a cluster by the its scores with the centroids of all clusters. The new centroid vecotr of each cluster is then updated as the average of the current centroid vector and the averaged vector of all text vectors belonging to this particular cluster. 

We show the $k$-means algorithm for dual encoder architectures in Algorithm~\ref{alg:k-means-dual}. 
\begin{algorithm}[]
\SetAlgoLined
\KwData{documents $\mathcal{T}$, categories $\mathcal{C}$, encoder $\enc$, scoring function $\mathrm{score}$
% $enc_c$ and $enc_t$
% , text index $k$, 
% \karl{Do we need to introduce a separate text index $k$ when we can just use the document identity $t$? Likewise, do we need to have category index $i$ when we have category identity $c$? If not it'd be much more clear to just say ``for document $t \in \mathcal{T}$ compute $s_{tc} = \mathrm{score}(enc_t(t), r_c)$ ...''}
% \zewei{Yeah I agree. It makes sense to simplify the notation. My concern is that we use $k$ and $i$ as score indices as well. If we change then to $t$ and $c$ will that cause some confusion? }
% category index $i$. 
}
 initialize the centroids as $r_c = \enc(c) \, \forall c \in \mathcal C$\;
 \While{not converged}{
    % \karl{Is this an implicit for loop over documents and categories, or is it for a single specific $(k, i)$? If former, either put the loop here or explain this implicit iteration in the caption.} 
    \For{$t \in \mathcal T$}{
        \For{$c \in \mathcal C$}{
            $s_{tc} = \mathrm{score}(\enc(t), r_c)$\;
        }
        $\mathrm{pred}_t = \mathrm{argmin}_c s_{tc}$\;
    }
    \For{$c \in \mathcal C$}{
        $r_c = (\frac{\sum_{t : \mathrm{pred}_t = c}\enc(t)}{\mathrm{count}(\{t : \mathrm{pred}_t = c\})} + \enc(c)) / 2$\;
    }
    $\mathrm{objective} = \sum_t s_{t \mathrm{pred}_t}$\;
 }
 \KwResult{$s_{tc}$, $\mathrm{objective}$, $\mathrm{pred}_t$, and $r_c$ of all iterations}
 \caption{Unsupervised label refinement for dual encoder architectures 
%  \kevin{In the update line, I think the $enc(c_i)$ should be $enc_c(c_i)$, right? Also, what is meant by $enc(s_k)$? I thought that should be something like $enc_t(t_k)$? Also, I'd suggest changing $ct$ to some single-letter notation ($r$?) because 2-letter symbol names can be confusing to people (it looks like a product).}
% \zewei{corrected}
}
 \label{alg:k-means-dual}
\end{algorithm}
We use one encoder $\enc$
% Here $enc_c$ and $enc_t$ are 
to encode texts and categories.
% encoders respectively. 
%We initialize the centroids using category encodings. 
We link centroids to categories and initialize the centroids using the encodings of the corresponding categories. 
%use our encoder to encode each category and provide its initial position.
%Since the distance between the category vector and the text vector is used to perform classification for the text, the category vectors can be naturally treated as the initial centroids for $k$-means clustering. 
The algorithm converges when no data point (text representation) updates its cluster assignment, i.e., the centroids stop updating. 
In our experiments, we run a maximum of 100 iterations. 
%in our $k$-means algorithm. 
We perform model selection (``early stopping'') based on the minimum value of $\mathrm{objective}$ among iterations. 
% Note that our $k$-means algorithm differs from the standard $k$-means algorithm as our updated centroid in each iteration is the average of the ``new centroid''(average of vectors in that cluster) and the old centroid. The reason behind it is that the standard $k$-means clustering algorithm is applied in an unsupervised setting, where the initial centroids do not have real meanings. While our initial centroids are encoded category vectors, 

Our $k$-means algorithm %in the dual encoder setting 
differs from standard $k$-means as our updated centroids are interpolated with the initial category embeddings. 
%The motivation behind this is that, unlike 
In standard $k$-means, the centroids are typically initialized randomly. 
%and do not have real meanings behind it. While 
In our case, since we link centroids to categories and use our encoder to provide initial centroids, we want to 
% continue to 
leverage the information in our category embeddings across clustering iterations. 
%have an informative encoding the centroids are initialized to vectors output by our encoder. 
Therefore, we average the ``new centroid'' with the original category vector, which serves as a kind of regularization. 
In preliminary experiments we found this modification to stabilize performance so we use it in our experiments reported below.

\subsection{\ulr for Single Encoder Architectures}

In our single encoder architecture, a score is produced for each document-category pair indicating its relatedness. For each document, after exponentiating and normalizing the score over all categories, we obtain a distribution indicating the probability of the document belonging to each category. For text $t$ and category $c$, we have a probability $p_{tc}$, where $\sum_c p_{tc} = 1$. 

A straightforward way of classifying each document is to pick the category of which it has the highest probability score, i.e., $\mathrm{argmax}_c p_{tc}$. Another way of interpreting this classification rule is to define $\vert \mathcal{C} \vert$ centroid vectors, each being an identity distribution $\mathrm{one \mbox{\textendash} hot}(c)$,\footnote{Here we abuse the notation of $\mathrm{one \mbox{\textendash} hot}(c)$ to represent a one-hot vector that has a ``1'' at the index of category $c \in \mathcal{C}$.} and pick the category having the minimum Jensen-Shannon Divergence \cite{lin91js} with the document probability vector, i.e., $\mathrm{argmin}_c \mathrm{JS}(p_{t}, \mathrm{one\mbox{\textendash} hot}(c))$. 
%\kevin{shouldn't it be a min over JS instead of a max? alg 2 shows min} \zewei{corrected} 
Here $\mathrm{JS}$ is the function to compute the Jensen-Shannon Divergence between two distributions. 
%\kevin{Hmm. Jensen Shannon Divergence is indeed the average of two KLs, but the two KLs use the average of the two distributions (see wikipedia). Zewei, in your experiments, did you use the real Jensen Shannon divergence or the function below?} \zewei{I used the function provided by sklearn \url{https://scipy.github.io/devdocs/generated/scipy.spatial.distance.jensenshannon.html}, so it is the right implementation. I updated the equations below. }
% \begin{align}
%     \mathrm{JS}(p, q) &= \frac{D_{\mathrm{KL}}(p, m) + D_{\mathrm{KL}}(q, m)}{2} \\
%     D_{\mathrm{KL}}(p, m) &= \sum_i p_i \log(\frac{p_i}{m_i}) \\
%     m &= (p + q) / 2
% \end{align}
Some prior work has explored using probability distribution based distance metric for clustering~\citep{banerjee2005clustering}. 
% \zewei{I put a single sentence here. Talk about k-means on probability distribution. \url{https://www.jmlr.org/papers/volume6/banerjee05b/banerjee05b.pdf}}

\begin{algorithm}[]
\SetAlgoLined
\KwData{documents $\mathcal{T}$, categories $\mathcal{C}$, scoring function $\mathrm{score}$, function to compute JS divergence $\mathrm{JS}$
% \karl{There's an inconsistency between inputs for Algorithm 1 and inputs for Algorithm 2 (absence of text/category index). Make them as consistent as possible for clarity. Same comments about the index notation and an implicit for loop.}
}
 initialize the centroids as $r_c = \mathrm{one \mbox{\textendash} hot}(c) \, \forall c \in C$\;
 %\zewei{Is this notation for one-hot vector correct, as $c$ is a category, but not an index? }\kevin{I think it's fine given that we now define the one-hot function in footnote 3}
 compute each document's probability distribution over categories as $[p_t]_c \propto \exp(\mathrm{score}(t,c)) \,\, \forall t\in\mathcal{T}, c \in \mathcal{C}$\;
 %\karl{Write $[P_t]_c \propto \exp(\mathrm{score}(t,c)) \,\, \forall t\in\mathcal{T}, c \in \mathcal{C}$}\kevin{agreed} \zewei{updated} \;
 \While{not converged}{
    \For{$t \in \mathcal{T}$}{
        \For{$c \in \mathcal{C}$}{
            $s_{tc} = \mathrm{JS}(p_t, r_c)$\;
        }
        $\mathrm{pred}_t = \mathrm{argmin}_c s_{tc}$\;
    }
    \For{$c \in \mathcal{C}$}{
        % Update the centroids as 
        $r_c = \frac{\sum_{t : \mathrm{pred}_t = c}p_t}{\mathrm{count}(\{t : \mathrm{pred}_t = c\})} $\;
    }
    
 }
 \KwResult{$\mathrm{pred}_t$ of the last iteration}
 \caption{$k$-means algorithm for single encoder architectures}
 \label{alg:k-means-single}
\end{algorithm}

It is natural to represent each category as a distribution by a one-hot vector. However, in a real text classification problem, the semantics of a category is affected by how the annotators view it. For instance, a news document could relate to both ``business'' and ``science \& technology'', though it will only have a single annotated category in the downstream task dataset. With this intuition, we propose to represent each category by a soft distribution over all the categories, but not necessarily a one-hot vector. 

% \zewei{Try KL divergence rather than JS divergence. Two orders, c/P or P/c?}

Algorithm~\ref{alg:k-means-single} describes our $k$-means clustering approach applied on the single encoder model. In this algorithm our predicted categories will be the clustering assignments of the last iteration. 
%\kevin{So there's no early stopping with this method? That would be good to clarify, e.g., ``Unlike with the dual encoder model, we do not do early stopping.''} \zewei{added}
Unlike with the dual encoder model, we do not do early stopping. 
% \karl{Also, maybe explain why we are not regularizing by interpolating with the one-hot initial label embeddings in this case? I guess it's because they are not as informative as the dense counterparts in dual encoder?}
We also do not use interpolated centroids as one-hot vectors may not necessarily be good category embeddings in this setting. 
% \zewei{Is this a good explanation? }
% \kevin{Sounds good to me for the submission. maybe by  camera-ready time we'll have a richer explanation}
% \zewei{Actually I found the performance drop slightly (about 1 point for these tasks), but I don't think we should put this as the reason? }

% \zewei{We are always using the category information in every step. }

\section{Experimental Setup}
\label{sec:experiments}

% \zewei{This is the first step of our XXXX framework. Our goal is to train an imperfect general purpose text classifier. }

% \zewei{Make this section to 25\% of the current length. }

% In this section, we describe the 

% Text classification aims at assigning given text to its most suitable category among a set of categories. For a particular text classification task, let the set of all possible classes be $\mathcal{C}$. Given a piece of text $t$, and a scoring function $\mathrm{score}$ that measures how likely the text belongs to a particular category $c \in \mathcal{C}$, we now have a score for each category $\mathrm{score}(c, t) \forall c \in C$. The text is categorized into the class with the highest score $\max_{c \in C}\mathrm{score}(c, t)$. 

% In this section, we introduce the models we used to build zero-shot text classifiers. We also introduce the datasets we used to train and evaluate such text classifiers. 

% \zewei{Standard dataless classification is based on text embeddings. Our experiments are based on freely available annotated text from \natcat. }

In this section, we introduce the datasets we used for evaluation and 
% datasets we use to train 
the dual encoder and single encoder dataless classifiers we used to run \ulr experiments. 
% and the text classification datasets for evaluation. We also describe the exact models we build to run the dual encoder and single encoder experiments. 

\subsection{Evaluation}

% \zewei{In this subsection, we discuss briefly introduce the datasets we used for training (\natcat). I could publish this dataset somewhere (GitHub or Google drive) and cite our dataset? }

%in the following text are trained on such binary classification tasks as of whether a document-category pair is accurate or not. 

% The \natcat dataset can be used to train as a binary classification task. 

% We formulate \natcat training as a binary classification task to predict whether a category correctly describes a document. 
% For each document-category pair, we randomly sample 7 negative categories for training. As documents from Wikipedia have multiple positive categories, we randomly sample one positive category for each of them. 

% Training on this binary classification task yields a scoring function which indicate the relatedness of a category and a piece of text. 

We use four text classification datasets spanning different domains for evaluation. They are: AG News\footnote{\url{https://www.di.unipi.it/~gulli/AG_corpus_of_news_articles.html}} (AG), which uses 4 classes and covers the newswire domain; 
DBpedia (DBP; \citealp{lehmann2015dbpedia}), which has 14 classes and is from the domain of encyclopedias;  
Yahoo \cite{zhang-char-conv}, which has 10 classes and addresses categorizing questions in online question fora; 
and 20 newsgroups~(20NG; \citealp{lang-95}), with 20 classes which are types of newsgroups. 
%\kevin{I added some more details about numbers of classes above -- Zewei, please check to make sure I got the numbers of classes correct.} \zewei{They are correct}
 
We do not use any data or labels from the training set in our main experiments, but only rely on label descriptions. 
We use the official label names from these datasets, and only expand them if the original label name is provided as abbreviations such as ``sci\_tech''.  The exact label names we used are in the appendix. 

\subsection{Dataless Classifiers}

% \zewei{In this section, we will describe the model architectures and loss functions we used in our experiments: the single encoder model and the dual encoder model. }

% In this work, we consider two types of models that combine a category with a piece of text and produce a score. 
We experiment with multiple dataless text classifiers that vary in terms of their complexity. Our simplest classifier uses an encoder that averages pretrained GloVe~\citep{pennington-glove} word embeddings. 
%There is no additional training of this model on any data resources after the . %\roberta-base~\citep{liu2019roberta} (110M parameters) 
%For both single encoder and dual encoder experiments, 
We also fine-tune a \roberta model \citep{roberta} in both single and dual encoder settings, using \roberta-base (110M parameters). 
We choose \roberta instead of \bert as \roberta outperforms \bert in a variety of text classification tasks~\citep{roberta}. 
% \zewei{Do we need to give any reasons why we do not use ESA? 
% We do not experiment with traditional dataless classifiers such as ESA~\citep{chang-dataless, gabrilovich-esa} as they require much more memory to compute and does not perform as well as deep neural models such as \roberta. }
% \kevin{Yes, would be good to say that the best performing method from the original dataless classification work is a dual encoder model based on ESA, but that we focus on GloVe and RoBERTa because their low dimensional representations are more amenable to refinement. It would also be good to report some results in the text explicitly comparing RoBERTa and ESA, in order to make it clear that RoBERTa is better than ESA. Maybe that specific comparison of numbers can go in a footnote.}
We do not run experiments with traditional dataless classifiers such as ESA~\citep{chang-dataless, gabrilovich-esa} as ESA vectors are of extremely high dimension, making it computationally difficult to apply \ulr. GloVe and \roberta produce lower dimensional vectors that are more computationally amenable to refinement. Also, we experimented with ESA and found that it does not perform as well as our \roberta based models.\footnote{As a comparison to the results of the \roberta dual encoder in Table~\ref{tab:k-means-dual-results}, ESA accuracies (\%) are 71.2 for AG, 62.5 for DBP, 29.7 for Yahoo, and 25.1 for 20NG.}   
Next we describe the details of the three dataless classifiers that we use in our experiments. 

\paragraph{GloVe Dual Encoder.}
We use GloVe~\citep{pennington-glove} in the dual encoder setting, simply averaging word vectors to represent both the categories and the documents. 
We use the 300 dimensional GloVe vectors\footnote{\url{http://nlp.stanford.edu/data/glove.840B.300d.zip}} trained on Common Crawl.\footnote{\url{https://commoncrawl.org/}} 
We experiment with two  distance functions when using GloVe: cosine and L2. 

\paragraph{\roberta Dual Encoder.} 
%With the dual encoder model, 
The category $c$ and text $t$ are fed separately to \roberta using the formatting ``[CLS] $c$ [SEP]'' and ``[CLS] $t$ [SEP]''. 
%, and two vector representations are produced for them respectively. 
% The parameters of the category and text encoders are shared, which worked better than separating parameters in  preliminary experiments. 
%show that it gives better results than having two separate encoders. 
We use the average of the final-layer hidden states produced by \roberta as category and text vectors. 
A scoring function takes both the category and text vectors to produce a scalar value. 
%\kevin{which part of the encoder is used as input to the scoring function? in the single encoder model, it was CLS plus a linear transform and nonlinear activation.. is it the same here?} \zewei{I added some descriptions before this}
%In our experiments, 
We experiment with dot product and L2 distance as  scoring functions.

\paragraph{\roberta Single Encoder.} 
% With the single encoder model, the category is combined with the text as a whole and fed into an encoder model. The output of the encoder is a single vector that contains the information from both the category and the text. 

% With models such as \cite{devlin2019bert} and \cite{liu2019roberta},  
%During training, 
The category $c$ and text $t$ %from \natcat 
are combined in the form ``[CLS] $c$ [SEP] $t$ [SEP]'', tokenized, and encoded using \roberta. 
%In our experiments, 
We truncate $t$ to ensure the category-document pair is within 128 tokens. 
The vector representation of the ``[CLS]'' token (after a linear transformation and non-linear activation) is then passed to a linear layer to produce a score.

\subsection{Fine-tuning \roberta models}
We use the \natcat dataset
%\footnote{sample data will be provided in supplementary material} 
(Anonymous, 2020) 
% \zewei{(Anonymous, 2020)} 
to fine-tune the \roberta models. \natcat comprises document-category pairs from three resources: Wikipedia, Stack Exchange, and Reddit. Wikipedia documents are paired with their annotated categories and ancestor categories. Stack Exchange question and question descriptions are paired with their corresponding question domain. Reddit post titles are paired with their subreddit names. Additional details and sample instances from \natcat are provided in the supplementary material. 
Each \natcat document comes with positive and negative categories. A positive category describes the document, and a negative category is randomly sampled and is irrelevant to the document. The \roberta models are fine-tuned as binary classifiers to indicate whether a category is positive for a document. 
% We finetune a RoBERTa-base model (110M parameters) on the \natcat dataset to get a dataless text classifier.

We use the Huggingface framework~\citep{Wolf2019HuggingFacesTS} to fine-tune all \roberta models, using 
%The \roberta models are fine-tuned on 
300k instances from \natcat. 
% For \natcat combining three domains, we train on 300k instances. 
The peak learning rate is set to be 0.00002, and we perform learning rate warmup for 10\% of the training steps and then linearly decay the learning rate. 
We set the random seed to be 1 in all experiments. 
%We fix the random seed to 1 in all experiments. 
% As \roberta models are known to suffer from randomness among different runs, we perform each single experiment 5 times under different random seeds %(1, 11, 21, 31, 41) 
% and report the median of such five runs. 
% Our training code is built on Huggingface Transformers~\cite{Wolf2019HuggingFacesTS} and
% . Our training code 
% will be released upon publication. 

% \subsection{Predication}
% When applying the \natcat fine-tuned \roberta model on a particular text classification task with a fixed set of labels, the task can be performed as a ranking problem. More specifically, given some text to be categorized, we score the text with all potential categories. The text is classified into the category with the highest score among them. 

% \paragraph{Datasets for Training.} 

\paragraph{Fine-tuning \roberta dual encoder.}
% \kevin{I think readers will wonder why we pair dot product with binary cross entropy, and pair Euclidean with contrastive hinge. What's the reason?} \zewei{I feel cross entropy loss is always a good choice for binary classification tasks. However, with Euclidean distance, using cross entropy seems not possible, as distance is non-negative, so I picked contrastive hinge. Do we need to cite some papers with similar choices of loss functions to justify our choice? }\kevin{we don't necessarily need to cite papers, but we should justify it i think. i don't see any reason why euclidean can't be used with cross entropy.. the $\sigma$ function converts any scalar into the probability of true. we would just need to use negative euclidean distance instead of euclidean distance, but we would have to do that anyway with the contrastive hinge too, right? (see my later comment about negative euclidean distance with hinge)}
While many other combinations of score and loss function could be considered, we report results with two particular combinations: dot product paired with binary cross entropy and L2 distance paired with a contrastive hinge loss.
When dot product is used, 
\roberta is fine-tuned to minimize binary cross entropy between $\mathrm{score}(\enc(c), \enc(t))$ and a binary label $y \in \{0, 1\}$. 
% we use dot product as the scoring function for the dual encoder architecture, 
% \roberta is fine-tuned to minimize binary cross entropy:
% \begin{align}
%   x &= \mathrm{score}(\enc(c), \enc(t)) \\
%     \mathrm{loss} &= y \log \sigma(x) + (1 - y) \log (1 - \sigma(x))
% \end{align}
% where $\enc$ is the \roberta encoder, $\sigma$ is the logistic sigmoid function, and $y \in \{0, 1\}$ is a binary label indicating whether $c$ is a positive category for $t$. 

When using L2 distance, 
% as the scoring function in the dual encoder architecture, 
we fine-tune \roberta to minimize a contrastive hinge loss: 
%\zewei{What is the exact name of this loss?}\kevin{there are a bunch of names.. i tend to call this either a ``pairwise ranking hinge loss'' or a ``contrastive hinge loss''} \zewei{updated}
\begin{align}
x_p &= \mathrm{score}(\enc(c_p(t)), \enc(t)) \\
x_n &= \mathrm{score}(\enc(c_n(t)), \enc(t)) \\
\mathrm{loss} &= \max(x_p + \gamma - x_n, 0)
\end{align}
where $\mathrm{score}$ is the squared L2 distance, 
$c_p(t)$ is a positive category for text $t$,  $c_n(t)$ is a negative category, % that does not describe the text, 
and $\gamma$ is a parameter indicating the margin. 
%Intuitively speaking, 
This loss aims to make negative category-text pairs have higher squared L2 distance than negative pairs by the margin. 
% \kevin{in this case, is score actually negative Euclidean distance?} \zewei{I actually used squared Euclidean distance in my code. I think it is not negative. }\kevin{hmm, but then minimizing the loss leads to decreasing negative $x_p$, which corresponds to trying to increase the squared distance between the positive category and the text. that seems like the reverse of what we want.} \zewei{Ah you are right. I reversed the order of $x_p$ and $x_n$ in the equation. }
% \kevin{ok great!}

\paragraph{Fine-tuning \roberta single encoder.} 
When used as a single encoder, \roberta is fine-tuned to minimize binary cross entropy between the predicted score $\mathrm{score}(\enc(c, t))$ and a binary label $y$, 
% :
% \begin{align}
%   x &= \mathrm{score}(\enc(c, t)) \\
%     \mathrm{loss} &= y \log \sigma(x) + (1 - y) \log (1 - \sigma(x))
% \end{align}
where 
%$c$ is the category, $t$ is the text,  
%to produce a vector representation,  
$\mathrm{score}$ is a linear function that transforms the vector into a scalar score. 
%$\sigma(x) = \frac{1}{1 + \exp(-x)}$ is the sigmoid function, 
%correctly describes the text. % or not. 

\section{Experimental Results of ULR}

\paragraph{Dual Encoder Models.}

\begin{table}[t]
\setlength{\tabcolsep}{5pt}
\small
\begin{center}
\begin{tabular}{c|c|cccc|c}
\toprule
& & AG & DBP & Yahoo & 20NG & avg \\
\midrule
\multirow{2}{*}{ cosine } & baseline & 66.1 & 43.5 & \bf 29.4 & 29.8 & 42.2 \\
& + \ulr & \bf 78.3 & \bf 46.4 & 27.9 & \bf 30.8 & \bf 45.9 \\
\midrule
\multirow{2}{*}{ L2 } & baseline & 40.5 & 15.7 & 14.5 & 19.5 & 22.6 \\
& + \ulr & \bf 58.7 & \bf 35.7 & \bf 23.3 & \bf 25.9 & \bf 35.9 \\
\bottomrule
\end{tabular}
\end{center}
\caption{\label{tab:k-means-glove-dual-results} 
Accuracies (\%) when applying ULR to the GloVe dual encoder architecture (``baseline''), for two different choices of distance function (cosine and L2).}
%Applying K-means with the GloVe word vectors. Accuracy on each task. }
\end{table}

With the dual encoder models, we used two sets of distance measures. The first distance is the cosine distance of two vectors. In this case, we always normalize the vector representations before applying ULR. The second distance measure is the L2 distance.

Table~\ref{tab:k-means-glove-dual-results} shows results for the GloVe dual encoder model with two 
% different 
distance functions. Except for the single case of cosine distance with  Yahoo, all accuracies improve, with some improving by large amounts (up to 20\% absolute). 

\begin{table}[t]
\setlength{\tabcolsep}{5pt}
\small
\begin{center}
\begin{tabular}{c|c|cccc|c}
\toprule
& & AG & DBP & Yahoo & 20NG & avg \\
\midrule
\multirow{2}{*}{ cosine } & baseline & \bf 74.0 & 84.6 & 52.3 & 36.4 & 61.8 \\
& + \ulr & 73.7 & \bf 93.3 & \bf 54.3 & \bf 40.3 & \bf 65.4 \\
\midrule
\multirow{2}{*}{ L2 } & baseline & 69.9 & 78.8 & 55.7 & \bf 37.8 & 60.6 \\
& + \ulr & \bf 70.7 & \bf 90.5 & \bf 61.3 & 37.4 & \bf 65.0 \\
\bottomrule
\end{tabular}
\end{center}
\caption{\label{tab:k-means-dual-results} Accuracies (\%) when applying ULR to the \roberta dual encoder  architecture (``baseline'').}
\end{table}

With \roberta dual encoder model, the choice of distance function matches the scoring function we used when fine-tuning \roberta, i.e., cosine distance is used with dot product scoring and L2 distance is used with the contrastive hinge loss. 
% \kevin{I think we need to make sure we clarify the score/distance function choices. We described two different distances in the dual encoder model description above (dot and Euclidean), and also described a score function in Alg 1 -- do we always use the same score function in Alg 1 that we use in the dual encoder model?} \zewei{I have updated the paragraph above. }
% When performing $k$-means clustering in label refinement, we did early stopping on the predicted scores of the average predicted categories \zewei{add more details, or an equation}.  
Table~\ref{tab:k-means-dual-results} shows results of ULR in the dual encoder setting with \roberta encoders. ULR improves accuracies by 3.6\% to 4.4\% on average, and the improvements are consistent across distance functions and datasets, except for the cases of \twentynews with L2 distance and AG with cosine distance, which show slight degradations. 
%In almost all text classification tasks with both cosine and L2 as the distance metrics, our $k$-means clustering approach improves performance, except in \twentynews with L2 loss, the performance drops slightly by 0.4 points. 

\paragraph{Single Encoder Model.}
\begin{table}[t]
\setlength{\tabcolsep}{5pt}
\small
\begin{center}
\begin{tabular}{c|cccc|c}
\toprule
 & AG & DBP & Yahoo & 20NG & avg \\
\midrule
baseline & 72.6 & 81.8 & 59.3 & 36.0 & 62.4 \\
+ \ulr & \bf 75.1 & \bf 88.6 & \bf 60.0 & \bf 36.5 & \bf 65.1 \\
\bottomrule
\end{tabular}
\end{center}
\caption{\label{tab:k-means-single-results}
Accuracies (\%) when applying ULR to the \roberta single encoder  architecture (``baseline'').  
% Applying $k$-means clustering with the single encoder architecture. The distance metric is Jensen-Shannon Divergence. Accuracy on each task. 
}
\end{table}

Table~\ref{tab:k-means-single-results} summarizes the results of applying \ulr to the \roberta single encoder architecture. \ulr improves performance across all four datasets, ranging from 0.5\% for 20NG up to 6.8\% on DBP. 

\paragraph{Label Ensembles.}
Finding the best choice of label names for dataless classifiers can be difficult without labeled data. Therefore, it is easier to supply multiple choices of label names for a given task.

We manually pick 10 different sets of label names for a particular task, generate category and text representations with our \roberta dual encoder model, and perform \ulr. The exact choices of label names are in the appendix. The predicted scores of such 10 different settings are summed up as the final ensemble predictions. Table~\ref{tab:k-means-dual-ensemble-results} presents the results on such ensemble predictions. Compared to Table~\ref{tab:k-means-dual-results}, all accuracies on all tasks are improved. Even with a stronger starting point from ensembling, \ulr still yields consistent improvements in accuracy, with the single case of Yahoo and cosine distance being the only one that does not improve. 

\begin{table}[t]
\setlength{\tabcolsep}{6pt}
\small
\begin{center}
\begin{tabular}{c|c|ccc}
\toprule
 & & AG & DBP & Yahoo  \\
 \midrule
\multirow{2}{*}{ cosine } & ensemble & 79.1 & 84.6 & \bf 54.5 \\
& + \ulr & \bf 81.1 & \bf 93.3 & 54.4 \\
\midrule
\multirow{2}{*}{ L2 } & ensemble & 72.4 & 79.0 & 56.5  \\
& + \ulr & \bf 73.4 & \bf 90.7 & \bf 61.7  \\
\bottomrule
\end{tabular}
\end{center}
\caption{\label{tab:k-means-dual-ensemble-results} Accuracies (\%) of ensemble predictions using 10 choices of label names using the \roberta dual encoder architecture (``ensemble''), and results when combining it with \ulr. 
% \zewei{How far are we from the optimal label choices. Put this in text. }
}
\end{table}

\section{Robustness to label noise and random label initialization}

% \zewei{This would make an interesting section. We use different label names for the same set of categories and see how that affect the final results in our framework. Another setting is to start from random centroids and see how K-means could yield good clusters. }

% \zewei{We could also visualize how the centroids evolve in K-means in this section. }

% \zewei{We could also ask the users to provide different label names and we ensemble them. }

% \zewei{Give some examples showing what label sets are good and what label sets are not good. Look at the label names. Think about what other points we want to make? }

% \zewei{Show some plots showing the starting accuracy and the improvement. X axis is the starting accuracy, y accuracy is the improvement by perentage. }

\paragraph{Different choices of label names.} 
%\kevin{I think we should be careful about the use of the term ``random'' here, because we actually mostly use label descriptions that are semantically similar to the original labels as we understood them. That is, we didn't do any truly random choices in this subsection's results.} \zewei{I changed it to ``different'', does it sound better? }\kevin{yes, sounds good!}
Dataless text classification tasks are known to suffer from high variance due to label descriptions. Performance can vary dramatically across different choices of category names.  Table~\ref{tab:example-noisy-labels} shows some examples of how different choices of labels names drastically change the performance of our \roberta dual encoder model fine-tuned with dot product scoring function. 
% \zewei{Show some experiment results of how changing label names affect the model performances? }

One advantage of \ulr 
%the K-means algorithm with dual encoder models for zero-shot text classification 
is that it is robust to label noise. That is, even given a poor choice of label names, \ulr can help the classifier to partially recover some of the accuracy, as shown in the lower portion of Table~\ref{tab:example-noisy-labels}. 

\begin{table}[t]
\setlength{\tabcolsep}{5pt}
\small
\begin{center}
\begin{tabular}{c|c|ccc}
\toprule
\multicolumn{2}{c|}{} 
& AG News & DBP & Yahoo \\
\midrule
%\multirow{2}{*}{cos.} 
& baseline & 69.6 & 79.4 & 43.8 \\
cos. & + \ulr & \bf 77.5 & \bf 91.6 & \bf 45.3 \\
& \# imp. & $\frac{\textrm{213}}{\textrm{240}}$ = 88.8\%  & $\frac{\textrm{3867}}{\textrm{3867}}$ = 100\% & $\frac{\textrm{732}}{\textrm{1015}}$ = 72.1\% \\
\midrule
& baseline & 62.2 & 71.7 & 49.5 \\
L2 & + \ulr & \bf 75.2 & \bf 85.2 & \bf 59.3 \\
& \# imp. & $\frac{\textrm{237}}{\textrm{240}}$ = 98.8\% & $\frac{\textrm{2960}}{\textrm{2963}}$ = 99.9\% & $\frac{\textrm{947}}{\textrm{947}}$ = 100\% \\
\bottomrule
\end{tabular}
\end{center}
\caption{\label{tab:robust} Robustness analysis when varying choices of label names. ``baseline'' and ``\ulr'' are average accuracy (\%) of the \roberta dual encoder architecture among all category naming choices. ``\# imp.'' are the numbers and percentages of cases where the performance improves after \ulr. }
\end{table}

\begin{figure}[t]
    \centering
    \includegraphics[scale=0.19]{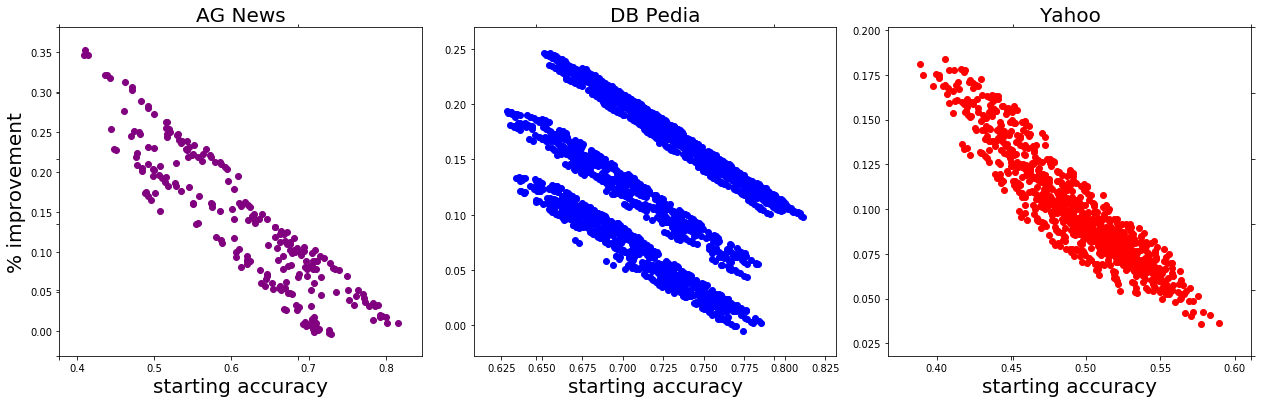}
    \caption{ Accuracy (\%) improvement  
    % Relative gains 
    when applying ULR compared to the \roberta dual encoder model (with Euclidean distance). The horizontal axis is the initial accuracy and the vertical axis is the absolute accuracy improvement after ULR. 
    }
    \label{fig:robust_improve}
\end{figure}

To demonstrate that 
%our $k$-means clustering based label refinement approach 
\ulr for dataless text classification is robust to label noise, we now describe similar experiments on a larger scale. In particular, we test several hundred sets of label names for each of 
%with applying different label names for the same set of fixed categories in the tasks of 
\agnews, \dbp, and \yahoo. For each category, we randomly assign different label names to it, all having similar meaning. The exact combinations of label names are in the appendix. 

We then perform dataless text classification and \ulr 
%$k$-means clustering 
with our \roberta dual encoder model, either under cosine distance or L2 distance. Table~\ref{tab:robust} shows the results of the average performance gains before and after \ulr. We also report the number and percentage of cases in which \ulr improves accuracy. 
%Our $k$-means clustering based label refinement approach 
\ulr improves the performance on average for all three tasks, and improves individual accuracies in the vast majority of cases. These results show that \ulr is not only effective at improving the accuracy of dataless text classifiers across a wide range of label name sets, but also can help to mitigate harmful effects due to suboptimal label names. 
%is also robust to a range of choices of label names. 

We next study the relationship between the \roberta dual encoder model's initial accuracies and its accuracy gains after applying \ulr. 
Figure~\ref{fig:robust_improve} plots accuracy improvements vs.~initial accuracies for the three datasets. 
%In these plots the horizontal axes indicate the models' initial accuracies, and the y axes indicate how much  improvement ULR provides. 
We find that ULR gives larger gains when the initial accuracies are lower. 

%\zewei{I added the following paragraph }

Among all the combinations of label names that we tried, we also report the oracle accuracies corresponding to the best combinations we found (without ULR). With cosine distance, they are 81.9\% for \agnews, 87.5\% for \dbp, and 53.3\% for \yahoo. With L2 distance, they are 81.5\% for \agnews, 81.1\% for \dbp, and 58.9\% for \yahoo. Directly applying ULR with the \roberta dual encoder classifier provides better performance on \dbp and \yahoo, and competitive results on \agnews, as shown in Tables \ref{tab:k-means-dual-results} and \ref{tab:k-means-dual-ensemble-results}. 

\paragraph{Clustering with random initialization.}

We also investigate the impact of our initialization in \ulr. We consider a variation in which we randomly initialize centroids. 
Since we can no longer link clusters and categories, this task becomes unsupervised text classification. 
%We replace our category vectors from our encoder start with random initialization of category vectors (centroids). 
A standard $k$-means algorithm is applied to update the centroids. Unlike Algorithm~\ref{alg:k-means-dual}, in this experiment we do not interpolate the updated centroids with the initial centroids, as the initial centroids are randomly generated and do not have any meaning. 
% A similar label refinement approach as described in Algorithm~\ref{alg:k-means-dual} is applied to update the centroids. 
The accuracy is calculated based on the oracle one-to-one mapping between final centroids and categories. This is often referred to as ``one-to-one accuracy''. 
%in the unsupervised learning literature. 

% \begin{table}[t]
% \setlength{\tabcolsep}{5pt}
% \small
% \begin{center}
% \begin{tabular}{c|c|c}
% \toprule
% & & AG \\
% \midrule
% & baseline & 37.3 \\
% cosine & + ULR & 61.4 \\
% & \# improved & $\frac{\textrm{239}}{\textrm{240}}$ = 99.6\% \\
% \midrule
% & baseline & 39.9 \\
% L2 & + ULR & 75.8 \\
% & \# improved & $\frac{\textrm{240}}{\textrm{240}}$ = 100\% \\
% \bottomrule
% \end{tabular}
% \end{center}
% \caption{\label{tab:robust_random_centroids} Average 1-to-1 accuracy (\%) of \ulr with random initialization of centroids. ``baseline'' and ``\ulr'' are average accuracy (\%) of the \roberta dual encoder architecture among all random initialized centroids. ``\# imp.'' are the numbers and percentages of cases where the performance improves after \ulr.
% % Robustness analysis of unsupervised label refinement initialized from random centroids. 
% % \kevin{Are these average accuracies?}\zewei{Yes} 
% }
% \end{table}

\begin{table}[t]
\setlength{\tabcolsep}{5pt}
\small
\begin{center}
\begin{tabular}{c|c|cc}
\toprule
& & cosine & L2 \\
\midrule
& baseline & 37.3 & 39.9 \\
AG News & + ULR & \bf 61.4 & \bf 75.8 \\
& \# improved & $\frac{\textrm{239}}{\textrm{240}}$ = 99.6\% & $\frac{\textrm{240}}{\textrm{240}}$ = 100\%  \\
% \midrule
% & baseline \\
% L2 & + ULR \\
% & \# improved \\
\bottomrule
\end{tabular}
\end{center}
\caption{\label{tab:robust_random_centroids} Average 1-to-1 accuracy (\%) of \ulr with random initialization of centroids. ``baseline'' and ``\ulr'' are average accuracy (\%) of the \roberta dual encoder architecture among all random initialized centroids. ``\# imp.'' are the numbers and percentages of cases where the performance improves after \ulr.
% Robustness analysis of unsupervised label refinement initialized from random centroids. 
% \kevin{Are these average accuracies?}\zewei{Yes} 
}
\end{table}

Table~\ref{tab:robust_random_centroids} presents results on  \agnews with this experiment. 
We performed 240 trials with different random initialization of the centroids, and 
%our $k$-means based label refinement approach 
unsurprisingly, accuracy is improved in all cases.  
In this unsupervised setting, since we always choose the mapping between centroids and categories which maximizes accuracy, the initial (``baseline'' in Table~\ref{tab:robust_random_centroids}) centroid-category mapping may be different from the end (``+ ULR'') mapping. 
% \zewei{I added the following sentence. }
With L2 distance, \ulr with random initialization of centroids even outperform centroids initialized as category embeddings, which shows that unsupervised clustering is powerful in text classification with good text representations. 
However, we do not have the one-to-one mapping between category and final centroids in this pure unsupervised setting, so the results in Table~\ref{tab:robust_random_centroids} and Table~\ref{tab:k-means-dual-results}, \ref{tab:k-means-dual-ensemble-results} are not directly comparable. 

% \kevin{I think it's noteworthy that the L2 model gets a very high accuracy, 74.9\%, which is higher than the L2 accuracies in Tables 3 and 5, so reviewers may wonder about that. I wonder why that is the case actually. But I guess it would depend on what the variance is. Is the 74.9\% an average over the 20 random initializations? Do you know the standard deviation?}\zewei{standard deviation for random centroids is 0.09, and with our picked sets of label names is 0.04, so it does have higher variance. I will run for more rounds on this setting to see if we will have different results. If not I will just add a row for standard deviation in this table and table 6.   }
% \zewei{I have run some extra experiments for AG News with random starting centroids. However, it does look like ULR with random centroids work very well, i.e., unsupervised clustering works very well for \roberta dual encoder. In terms of standard deviation, with cosine, it is 0.04 for ``baseline'' and 0.07 for ``+ \ulr'', with L2, it is 0.05 for ``baseline'' and 0.08 for ``+\ulr''.  Compared to the standard deviation in table 6, they are: with cosine, it is 0.09 for ``baseline'' and 0.06 for ``+ \ulr'', with L2, it is 0.1 for ``baseline'' and 0.04 for ``+\ulr''.}

\section{Unsupervised Label Refinement with Few Shot Learning}

% \zewei{
% This section talks about how we can add labeled instances (30 labeled instances per category, same setting as the MSR paper). 
% The risk of this section is that we do not beat their performance on these tasks. The reviewers may think this is ``negative results'' as well. }

In this section, we apply \ulr in the few shot learning setting. 
In particular, we draw 30 labeled instances for each category from the training split in the original datasets. We then further fine-tune our \roberta dual encoder model with the labeled instances. 
We adopt the hyperparameters of fine-tuning text classifiers from \citet{Wolf2019HuggingFacesTS} and fine-tune for 3 epochs on these 30 labeled instances for each category. 
Then we apply ULR on the unlabeled test set in addition to the 30 labeled instances from each category. 

Unlike 
%Different from pure dataless $k$-means clustering as described in 
Algorithm~\ref{alg:k-means-dual}, 
% now we have 30 labeled instances for each category. 
these 30 labeled instances from each category are fixed to be assigned to their corresponding clusters. 
At every iteration, the update rule of the centroids will include the text vector representations of these 30 labeled instances. 
% \kevin{I'm a little confused by Alg 3. In the $r_c = $ update line, which $r_{lc}$ is used?}
% \zewei{$r_c^l$ never gets updated among iterations, so it is a fixed vector. }
% \kevin{ Or maybe I didn't quite understand the $r_{lc}$ update line. Is $r_{lc}$ the average of $\enc(t_l)$ for all labeled documents that are labeled with category $c$?}
% \zewei{correct}
% \kevin{ I think I'm also confused as to whether $\mathcal{T}_{lc}$ contains all labeled documents or only the labeled documents labeled with category $c$. }
% \zewei{only the 30 documents labeled with category $c$}
% \kevin{If it only contains the labeled documents labeled with category $c$, then I think we should use some different notation that separates the $l$ and $c$, like maybe use $\mathcal{L}_{c}$ to denote the set of labeled documents with category $c$. Then I think we should also change $r_{lc}$ to something like $\bar{r}_c$,  $r_c^\ast$, or $r_c^l$.}

\begin{table}[t]
\setlength{\tabcolsep}{5pt}
\small
\begin{center}
\begin{tabular}{l|cccc|c}
\toprule
& \agnews & \dbp & \yahoo & 20NG & avg \\
\midrule
baseline & 74.0 & 84.6 & 52.3 & 36.4 & 61.8 \\
+ 30 labeled ins. & 81.5 & 97.1 & 66.0 & 58.1 & 75.7 \\
+ ULR & \bf 85.2 & \bf 97.4 & \bf 66.7 & \bf 58.2 & \bf 76.9 \\
\bottomrule
\end{tabular}
\end{center}
\caption{\label{tab:labeled-k-means} Accuracies (\%) when adding 30 labeled instances for each category with the \roberta dual encoder model (``baseline''). }
\end{table}

\begin{algorithm}[]
\SetAlgoLined
\KwData{categories $\mathcal{C}$, unlabeled documents $\mathcal{T}$, documents $\mathcal{L}_c$ labeled with $c \,\, \forall c  \in \mathcal{C}$, encoder $\enc$, scoring function $\mathrm{score}$}
 initialize the centroids as $r_c = \enc(c) \,\, \forall c \in C$\;
 compute the labeled document centroids as $r_c^l = \frac{\sum_{t \in \mathcal{L}_c}\enc(t)}{\vert \mathcal{L}_{c} \vert } \,\, \forall c \in \mathcal{C} $\;
 \While{not converged}{
    \For{$t \in \mathcal{T}$}{
        \For{$c \in \mathcal{C}$}{
            $s_{tc} = \mathrm{score}(\enc(t), \enc(c))$\;
        }
        $\mathrm{pred}_t = \mathrm{argmin}_c s_{tc}$\;
    }
    \For{$c \in \mathcal{C}$}{
        $r_c = (\frac{\sum_{t : \mathrm{pred}_t = c}\enc(t)}{\mathrm{count}(\{t : \mathrm{pred}_t = c\})} + r_c^l + 2\times \enc(c)) /4$\;
    }
    $\mathrm{objective} = \sum_t s_{t \mathrm{pred}_t}$\;
 }
 \KwResult{$s_{tc}$, $\mathrm{objective}$, $\mathrm{pred}_t$, and $r_c$ of all iterations}
 \caption{$k$-means algorithm for dual encoder models, with 30 labeled instances}
\end{algorithm}

Table~\ref{tab:labeled-k-means} summarizes the results of combining 30 labeled instances for each category and ULR. Unsurprisingly, adding labeled instances improves accuracy by a large margin. Even so, ULR yields an additional improvement of 3.7\% for \agnews, and also improves on the other datasets. 
%While ULR further improves such few-shot trained classifiers in all evaluation tasks. 

\section{Augmenting Categories and Text}

% \zewei{This section talks about how adding extra categories or unlabeled text in the prediction domain could influence the performance of our proposed XXXX framework. }

In this section, we ask the following questions: will more unlabeled text inputs from the evaluation task improve \ulr? Will adding more category names benefit \ulr? 

\begin{table}[t]
\setlength{\tabcolsep}{4pt}
\small
\begin{center}
\begin{tabular}{c|c|cccc|c}
\toprule
& & \agnews & \dbp & \yahoo & 20NG & avg \\
\midrule
\multirow{4}{*}{ cosine } & baseline & 74.0 & 89.4 & 53.5 & 37.0 & 63.5 \\
& ULR & 73.7 & \bf 93.3 & 54.3 & 40.3 & 65.4 \\
& + aug cats & \bf 74.9 & 92.6 & \bf 57.3 & \bf 41.6 & \bf 66.6 \\
& + aug text & 73.8 & 93.2 & 54.2 & 40.3 & 65.4 \\
\midrule
\multirow{4}{*}{ L2  } & baseline & 69.9 & 78.8 & 55.7 & 37.8 & 60.6 \\
& ULR & 70.7 & \bf 90.5 & \bf 61.3 & 37.4 & 65.0 \\
& + aug cats & \bf 71.9 & 90.4 & 59.4 & \bf 39.8 & \bf 65.4 \\
& + aug text & 70.8 & 90.4 & \bf 61.3 & 37.6 & 65.0 \\
\bottomrule
\end{tabular}
\end{center}
\caption{\label{tab:augment} Accuracies (\%) when adding augmented text or categories to ULR with the \roberta dual encoder architecture (``baseline''). 
% \kevin{3 numbers are bold for AG L2 -- i think it should just be the aug cats result, right? Also, for the DBP cosine results, the bold result, 92.6, is not the highest.}\zewei{corrected}
}
\end{table}

%\paragraph{Augmenting categories.}
% \zewei{Pick categories with close meanings? }
When applying \ulr with dataless text classifiers, more augmented categories can be added to be the centroids of new clusters. 
The goal
% semantic meaning 
of adding such augmented categories is to enrich the semantic space of all possible categories. For example, a document belonging to ``science \& technology'' may be further classified into ``physics'', ``math'', or other scientific subjects. Adding such augmented categories into Algorithm~\ref{alg:k-means-dual} makes the category embedding cover broader and finer semantic spaces. 
To accommodate augmented categories in Algorithm~\ref{alg:k-means-dual}, we add more centroids which are updated across iterations. However, at prediction time, we only predict within the categories of the original label set. 

%\paragraph{Augmenting text.}

% \zewei{Adding unlabeled training data into our method. }

We also consider adding extra unlabeled text inputs from the domain of the evaluation task. The assumption is that by adding extra text, the $k$-means clustering approach would gain more knowledge from the task domain, and the centroids would capture semantics of larger scale data. 

% \zewei{I think we could add an experiment to fix a small test set, and gradually add more augmented text to see whether the performance improves or not. }

\paragraph{Experiments and results.} 

Starting from our \roberta based dual encoder model as a baseline, we run \ulr with augmented categories and text. Our augmented categories are the names of the 30 most popular Stack Exchange sites~\cite{chu2020mqr}. These category names cover a broad range of topics. The exact list of augmented category names are in the appendix. As for augmented text, we pick at most 100k instances without labels from the training sets of the evaluation tasks. The experimental results are summarized in Table~\ref{tab:augment}. Augmented categories give the most improvement to \ulr. Adding augmented text to \ulr has little impact, indicating that having more text does not always help in unsupervised label refinement.

% \section{Visualization}

% \section{Entropy based self training}

% Self training is a useful technique in many machine learning tasks. As a training technique, self training is orthogonal to any other 

% We 

\section{Related Work}
% \zewei{Copied from the \natcat paper, needs to be rewritten}
% Dataless classification is sensitive to the choices of label names. 
% \citet{chang-dataless,song-dataless-hierarchical,wang-universal-2009} 
% uses \esa (ESA) \citep{gabrilovich-esa} as text representations. 
% In their work, they manually expand the label names of the 20 news groups dataset so they capture the semantics of each category. 

% Our work is closely related to the idea of zero-shot text classification~\cite{yin-etal-2018-zero, puri2019zero}. 
%\footnote{The phrases ``weakly-supervised'' and ``zero-shot'' are sometimes interchangeable.} 
% \citet{puri2019zero} build text classifiers by training language models on statements regarding text and corresponding categories.  
% \citet{yin2019benchmarking} solves zero-shot text classification with an entailment approach. 
% They expand the labels by their WordNet definitions. 
% For instance, ``sports'' is interpreted as ``an active diversion requiring physical exertion and competition''. 

%\zewei{Move up to section 2. Write more related work. }

% build models by directly mapping Wikipedia documents to their annotated categories. We take a step further to incorporate text from a variety of freely available text-category pairs from online resources, and build models with both discriminative training of \bert-like models~\citep{devlin-etal-2019-bert} and generative training with \gpt~\citep{radford-2019}. 

% \zewei{More related works on label refinement}

We discussed prior work on dataless and zero-shot text classification in Sections \ref{sec:intro} and \ref{sec:background}. We briefly introduce more related work in this section. 
\citet{song2015spectral} provide a text label
refinement algorithm to adjust the label set with noisy and missing labels. 
There is also a wealth of prior work in semi-supervised text classification: using unlabeled text to improve classification performance~\cite{nigam-text-2000, mukherjee2020uncertainty, xie2019unsupervised}. These methods typically 
learn generally useful text representations from a large corpus of unlabeled text and 
use them for a specific target task with limited supervision \cite{howard-universal-2018,devlin2018bert,roberta,Lan2020ALBERT,peters2018deep}. 
Finally, in supervised classification, label descriptions are also exploited to improve text classifiers~\citep{chai2020description, wang2019aspect, sun2019utilizing}.

% Finally, %we mention prior work on 
% supervised text classification %which 
% is a well studied problem. 
% A typical approach is to convert text into a vector representation (e.g., bag-of-\textit{n}-grams) and apply standard classifiers 
% %classification models such as naive Bayes and support vector machine 
% \citep{wang-topic-classification-2012,joachims-text}.  
% Recent work based on neural networks achieves state-of-the-art performance \cite{,kim-cnn-2014,zhang-char-conv,johnson-dpcnn-2017,tang-sentiment-2015,johnson-semi-2015,johnson-shallow-deep-2016}. 
% In particular, attention mechanisms and joint document-label embeddings have been shown to be useful~\cite{yang-hierarchical-2016,wang-joint-2018}. 

% \natcat. 

\section{Conclusion}

% We propose a clustering based approach to mitigate the label noise problem in zero-shot text classifications. 

In this paper, we have shown that our proposed $k$-means clustering based unsupervised label refinement is a simple but effective approach to improve the performance of dataless text classifiers. ULR can be applied in both single encoder or dual encoder architectures. 
This approach is robust against the choices of label names, making dataless text classification more useful for practitioners.

% Our experiments show that this approach can be combined with either single encoder or dual encoder architectures. It closes the gap between zero-shot text classifiers and human to be within 10 points. 

\bibliography{aaai}
% \bibliographystyle{aaai21}

% \appendix
% \input{appendix}

\end{document}

\maketitle

\section{Experimental Setup}
We list the hyperparameters we used to fine-tune our \roberta based dataless classification models in this section, so readers can reproduce our experimental results. The code with the scripts are also included in the supplementary material. 
In all fine-tuning tasks, we set the batch size to be 32. The max sequence length for \roberta is 128. The peak learning rate is 0.00002. We use a linear scheduler with warmup steps to be 10\% of the total fine-tuning steps. The random seeds of all experiments are set to be 1. 

We fine-tune \roberta models on random choices of single GPUs, including NVIDIA TITAN X, 1080Ti, or 2080 Ti. Most of the fine-tuning tasks can be finished within 8 hours. 

% The \natcat dataset we used to fine-tune \roberta based dataless classification models are included in the 

\section{Label names in the evaluation tasks}

In the main experiments (Section 5), we use the following label descriptions (separated by ``;'') for the downstream tasks. 
\begin{itemizesquish}
\item \agnews: world; sports; business; science technology
\item \dbp: company; educational institution; artist; athlete; politician; transportation; building; nature; village; animal; plant; album; film; written work
\item \yahoo: society culture; science mathematics; health; education reference; computers internet; sports; business finance; entertainment music; family relationships; politics government
\twentynews
\item atheist christian atheism god islamic; graphics image gif animation tiff; windows dos microsoft ms driver drivers card printer; bus pc motherboard bios board computer dos; mac apple powerbook; window motif xterm sun windows; sale offer shipping forsale sell price brand obo; car ford auto toyota honda nissan bmw; bike motorcycle yamaha; baseball ball hitter; hockey wings espn; encryption key crypto algorithm security; circuit electronics radio signal battery; doctor medical disease medicine patient; space orbit moon earth sky solar; christian god christ church bible jesus; gun fbi guns weapon compound; israel arab jews jewish muslim; gay homosexual sexual; christian morality jesus god religion horus
\end{itemizesquish}

\paragraph{Label ensemble}
In the experiment of ensembling labels, we use the following 10 sets of label name choices. 
For \agnews, 
\begin{itemizesquish}
\item world; sports; business; science technology
\item international; sports; business; science technology
\item world; sports; business; science and technology
\item international; sports; business; science and technology
\item world; sports; business and finance; science technology
\item international; sports; business and finance; science technology
\item world; sports; business and finance; science and technology
\item international; sports; business and finance; science and technology
\item world politics; sports; business; science technology
\item world politics; sports; business; science technology
\end{itemizesquish}

For \dbp,
\begin{itemizesquish}
\item company; educational institution; artist; athlete; politician; transportation; building; nature; village; animal; plant; album; film; written work
\item company; school; artist; athlete; politician; transportation; building; nature; village; animal; plant; album; film; written work
\item company; educational institution; artist; athlete; politician; transportation; architecture; nature; village; animal; plant; album; movie; written work
\item company; educational institution; artist; athlete; politician; transportation; building; nature; village; animal; plant; album; film; novel
\item company; educational institution; artist; athlete; politician; transportation; building; nature; village; animal; plant; album; film; article
\item company; educational institution; artist; athlete; politician; transportation; building; nature; village; animal; plant; collection; movie; written work
\item company; school; artist; athlete; politician; transportation; architecture; nature; village; animal; plant; album; movie; written work
\item company; educational institution; artist; athlete; politician; transportation; architecture; nature; village; animal; plant; album; movie; novel
\item company; school; artist; athlete; politician; transportation; architecture; nature; village; animal; plant; album; film; novel
\item company; educational institution; artist; athlete; official; transportation; building; nature; village; animal; plant; album; movie; article
\end{itemizesquish}

For \yahoo,
\begin{itemizesquish}
\item society culture; science mathematics; health; education reference; computers internet; sports; business finance; entertainment music; family relationships; politics government
\item society culture; scientific discipline; health; education reference; computers internet; sports; business finance; entertainment music; family relationships; politics government
\item society culture; science mathematics; health; learning teaching resource; computers internet; sports; business finance; entertainment music; family relationships; politics government
\item society culture; science mathematics; health; education reference; computers internet; sports; business finance; entertainment music; love home; politics government
\item society culture; science mathematics; health; education reference; information technology; sports; business finance; entertainment music; family relationships; politics government
\item society culture; scientific discipline; health; education reference; information technology; sports; business finance; entertainment music; family relationships; politics government
\item society culture; science mathematics; health; learning teaching resource; information technology; sports; business finance; entertainment music; family relationships; politics government
\item society culture; science mathematics; health; education reference; computers internet; sports; commerce; entertainment music; family relationships; politics government
\item society culture; science mathematics; health; education reference; information technology; sports; commerce; entertainment music; family relationships; politics government
\item society culture; science mathematics; health; learning teaching resource; computers internet; sports; commerce; entertainment music; family relationships; politics government
\end{itemizesquish}

\section{Robustness to label noises}
In the experiment of robustness to label noises (Section 6), we manually picked different choices of label names for each category of the downstream tasks. We list the choices of label names for each category below, separated by ``;''.

\agnews: 
\begin{itemizesquish}
\item world: world; world politics; world news; international; international news; 
\item sports: sports; health; health and sports; 
\item business: business; commerce; finance; business and finance; 
\item science technology: science; technology; science and technology; science technology
\end{itemizesquish}

\dbp: 
\begin{itemizesquish}
\item company: company; corporation;
\item educational institution: educational institution; school;
\item artist: artist; creator;
\item athlete: athlete; sportsman;
\item politician: politician; official;
\item transportation: transportation; 
\item building: building; architecture;
\item nature: nature;
\item village: village; suburb;
\item animal: animal; living thing;
\item plant: plant;
\item album; collection;
\item file: film; movie;
\item written work: written work; writing; novel; article
\end{itemizesquish}

\yahoo:
\begin{itemizesquish}
\item society culture: society culture; community ;
\item science mathematics: science mathematics; scientific discipline;
\item health: health; fitness;
\item education reference: education reference; learning teaching resource;
\item computers internet: computers internet; information technology;
\item sports: sports; athletics;
\item business finance: business finance; commerce;
\item entertainment music: entertainment music; fun songs;
\item family relationships: family relationships; love home;
\item politics government: politics government; policy regime regulation
\end{itemizesquish}

Different combinations of label name choices are used to generate category embeddings and perform \ulr. 

\section{Augmented Categories}
In the experiment of augmenting categories with \ulr (Section 8), we use the following augmented categories (separated by ``;''): math; gis; physics; codereview; stats; unix; english; tex; gaming; apple; scifi; drupal; ell; meta; electronics; travel; rpg; dba; magento; webapps; diy; wordpress; android; security; chemistry; webmasters; blender; softwareengineering; gamedev; academia.